\documentclass[conference]{IEEEtran}
\IEEEoverridecommandlockouts
\usepackage[sorting=none]{biblatex}
\usepackage{amsmath,amssymb,amsfonts}
\usepackage{algorithmic}
\usepackage{graphicx}
\usepackage{textcomp}
\usepackage{xcolor}
\def\BibTeX{{\rm B\kern-.05em{\sc i\kern-.025em b}\kern-.08em
    T\kern-.1667em\lower.7ex\hbox{E}\kern-.125emX}}

\usepackage{commath}
\usepackage{enumitem}

\usepackage[hidelinks]{hyperref}
\hypersetup{
    colorlinks=true,
    linkcolor=orange,
    filecolor=magenta,      
    urlcolor=orange,
    citecolor=orange,
}
    
\begin{document}

\title{APReL: A Library for Active\\Preference-based Reward Learning Algorithms\\
    \thanks{This work is funded by NSF grants \#1849952 and \#1941722, FLI grant RFP2-000, and DARPA.}
}

    \author{\IEEEauthorblockN{Erdem B\i y\i k}
    \IEEEauthorblockA{\textit{Electrical Engineering} \\
    \textit{Stanford University}\\
    ebiyik@stanford.edu}
    \and
    \IEEEauthorblockN{Aditi Talati}
    \IEEEauthorblockA{\textit{Computer Science} \\
    \textit{Stanford University}\\
    atalati@stanford.edu}
    \and
    \IEEEauthorblockN{Dorsa Sadigh}
    \IEEEauthorblockA{\textit{Computer Science \& Electrical Engineering} \\
    \textit{Stanford University}\\
    dorsa@cs.stanford.edu}
    }

\maketitle

\begin{abstract}
Reward learning is a fundamental problem in human-robot interaction to have robots that operate in alignment with what their human user wants. Many preference-based learning algorithms and active querying techniques have been proposed as a solution to this problem. In this paper, we present APReL, a library for active preference-based reward learning algorithms, which enable researchers and practitioners to experiment with the existing techniques and easily develop their own algorithms for various modules of the problem. APReL is available at \url{https://github.com/Stanford-ILIAD/APReL}.
\end{abstract}



\section{Introduction}
As robots enter our daily lives, we want them to act in ways that are aligned with our preferences and goals. Learning a reward function that captures human preferences about how a robot should operate is a fundamental robot learning problem that is the core of the algorithms discussed in this work.

There are a number of different information modalities that can be avenues for humans to convey their preferences to a robot. These include demonstrations \cite{abbeel2004apprenticeship,ziebart2008maximum}, physical corrections \cite{li2021learning,bajcsy2017learning}, observations~\cite{stadie2017third}, language instructions and narrations~\cite{arumugam2019grounding}, ratings \cite{chu2005gaussian,shah2020interactive}, comparisons and rankings \cite{sadigh2017active,christiano2017deep,brown2019extrapolating}, each of which has its own advantages and drawbacks. For example, demonstrations are difficult to collect when the robot has high degrees of freedom, and physically providing corrections on a robot can pose safety challenges. Observations are difficult to learn from, as they are provided by agents with different dynamics and/or objectives. Language instructions and narrations do not provide detailed information about how the task should be done. While ratings do not suffer from these issues, they introduce higher cognitive loads on the users compared to comparisons and rankings. Learning human preferences using comparisons and rankings is well-studied outside of robotics~\cite{de2009preference}, and the paradigm of learning human preferences based on comparisons and rankings shows promise in robotics applications as well~\cite{biyik2021learning,christiano2017deep,brown2019extrapolating}. 

However, preference-based learning poses another important challenge: each comparison or ranking gives a very small amount of information. For example, a pairwise comparison between a trajectory of a car that speeds up at an intersection with another trajectory that slows down gives at most one bit of information. Hence, it becomes critical to optimize for what the user should compare or rank. To this end, prior works have developed several active learning techniques to improve data-efficiency of preference-based learning by maximizing the information acquired from each query to the user. In this paper, we present a \emph{novel} and \emph{unified} Python library, APReL, that enables researchers and practitioners to easily use and experiment with many existing active preference-based reward learning techniques that effectively query humans and are used in robotics applications. Therefore, it is very \emph{relevant} to both human-robot interaction and robot learning fields. For the \emph{ease of use}, APReL enables these various techniques to be applied on any simulation environment that is compatible with the standard OpenAI Gym structure \cite{brockman2016openai}.

We first go over the techniques included in APReL in Section~\ref{sec:related_work}. We also discuss more recent studies that we are actively working to include in later versions of APReL. Next, Section~\ref{sec:problem} presents a unifying notation, and Section~\ref{sec:methods} briefly reviews the techniques. Section~\ref{sec:usage} presents the modular structure of APReL to guide the researchers for implementing new techniques. Finally, Section~\ref{sec:conclusion} concludes the paper.

\section{Related Work}\label{sec:related_work}

In this section, we go over prior work that APReL supports, and discuss more recent studies which we plan to include in APReL's later versions.

\smallskip
\noindent\textbf{Active Preference-Based Learning.} Active preference-based reward learning is a well-studied problem in machine learning and robotics. \textcite{sadigh2017active} modeled the reward as a linear function of some trajectory features and proposed using a \emph{volume removal} based acquisition function to select pairwise comparison queries, which are in the form of ``do you prefer trajectory A or B?". \textcite{katz2019learning} formulated the query selection as a multi-objective optimization problem to maximize both the likelihood and the \emph{disagreement} between trajectories.

While the disagreement-based method was only for pairwise comparisons, \textcite{palan2019learning} and \textcite{biyik2019green} extended the volume removal method to best-of-$K$ queries: ``which of the $K$ trajectories is the best?". Later, \textcite{biyik2019asking} found the volume removal optimization is ill-posed and identified failure cases. They proposed maximizing \emph{mutual information} between the response to the query and the reward function. Finally, \textcite{wilde2020active} and \textcite{tucker2020preference}, again focusing on pairwise comparisons only, developed alternative acquisition functions based on a regret formulation and Thompson sampling, respectively. These further improve the data-efficiency when the goal is to find the optimal trajectory rather than the underlying reward function. Although APReL focuses on reward learning, we include these acquisition functions in the library as additional benchmarks.

\smallskip
\noindent\textbf{Batch Active Learning.} While these works focus on the data-efficiency of learning, another line of work built upon them to improve time-efficiency by generating batches of queries at a time. This reduces the time taken to generate each query at the expense of increased number of queries needed to learn the reward function. When generating a batch, however, simply selecting the top few queries in terms of the acquisition functions will create a batch of nearly-identical queries. Most of these queries will be redundant after the user responds to one of them, and the data-efficiency will significantly decrease. To handle this problem, \textcite{biyik2018batch} adopted various heuristic methods to jointly maximize informativeness and diversity. They later developed a method based on Determinantal Point Processes (DPP) \cite{biyik2019batch} which are used for tractably sampling informative and diverse batches in a more formal way \cite{kulesza2012determinantal}. APReL enables both the heuristic methods and the DPP method in addition to the non-batch setting.

\smallskip
\noindent\textbf{Integrating Comparisons and Demonstrations.} Finally, recent works attempted to combine other sources of information with comparisons and rankings. \textcite{palan2019learning} and \textcite{biyik2021learning} integrated expert demonstrations into the prior to further improve data-efficiency. Using their methods based on Bayesian inverse reinforcement learning \cite{ramachandran2007bayesian}, APReL enables researchers to optionally include demonstrations (or other forms of offline data) into the learning pipeline and still use all other functionalities.

Although APReL covers many of the techniques in the literature, preference-based reward learning is still an active research field. Hence, there exist recent methods that we are actively working towards adding into the library. These include scale feedback questions \cite{wilde2021learning}; reward functions that are non-stationary \cite{basu2019active}, multimodal \cite{myers2021learning}, modeled as Gaussian processes \cite{biyik2020active} (which was later combined with ordinal data by \textcite{li2021roial}) or neural networks \cite{katz2021preference}; and a new acquisition function that enables the robot to reveal what it has already learned, while asking questions \cite{habibian2021here}.

\section{Problem Formulation}\label{sec:problem}
APReL synthesizes various techniques discussed in Section~\ref{sec:related_work}. To this end, we start with formulating the problem with a unifying notation.

\smallskip
\noindent\textbf{MDP.} We describe the evolution of the agent as a discrete-time Markov Decision Process (MDP) $\mathcal{M} = \langle\mathcal{S}, \mathcal{A}, f, r\rangle$. At each time step $t$, the agent is at state $s_t \in \mathcal{S}$ and takes action $a_t \in \mathcal{A}$. Then, $f$ represents the agent's dynamics distribution such that $s_{t + 1} \sim f(s_t, a_t)$. Finally, $r$ is the reward function, so that at every time step $t$, the agent receives a reward $r(s_t,a_t)$.
A trajectory $\xi \in \Xi$ in this MDP is a sequence $\del{\del{s_t, a_t}}_{t}$ of state-action pairs that correspond to a roll-out in the MDP. The goal of the MDP agent is to maximize the expected cumulative reward over its trajectories.

\smallskip
\noindent\textbf{Reward.} The reward function is what we are trying to learn, as it encodes how the human wants the agent to behave. We assume it is linear in some state-action features that are known: $r(s_t,a_t) = \omega^\top \phi(s_t,a_t)$.\footnote{This is a common assumption to make the Bayesian learning approaches tractable. We are actively working on extending these to more expressive reward models in APReL. See, for example, \cite{christiano2017deep,biyik2020active,katz2021preference}.} The reward of a trajectory $\xi$ is then based on the cumulative features:
\begin{align}
	R(\xi) =\! \sum_{(s,a) \in \xi}\! r(s,a) = \omega^\top\!\sum_{(s,a) \in \xi}\!\phi(s,a) = \omega^\top \Phi(\xi)\:.   
\label{eq:reward}
\end{align}
In fact, as we elicit user preferences by asking them to compare or rank trajectories, rather than states or actions, this formulation allows us to define the features more generally. As an example, one can directly design $\Phi$, which might treat states at different time steps differently, rather than designing $\phi$. Since $\Phi$ is given, we only have to learn the weights $\omega$.

\smallskip
\noindent\textbf{Preferences.} The human gives information by selecting their preference among a query $Q = \cbr{\xi_1, \xi_2, \ldots, \xi_K}$ of $K$ trajectories. The human noisily picks their favorite $q \in Q$, which optimizes their reward function. In addition, differently from the existing work, APReL also allows users to give full rankings of the trajectories in $Q$. We use these preferences and rankings to learn the human's reward function.

\smallskip
\noindent\textbf{Demonstrations.} Expert demonstrations of the optimal behavior may also be available to initialize the learning process. Each demonstration is a trajectory $\xi^D$, and the human may input a set of demonstrations as $\mathcal{D}$.

\smallskip
\noindent\textbf{Problem.} Our overall objective is to learn the reward function with as few data points as possible. For this, we should:
\begin{itemize}[nosep]
    \item Learn from user preferences after optionally initializing our model based on expert demonstrations,
    \item Actively generate preference/ranking queries that are optimized to be informative for the learning model, and
    \item Actively generate \emph{batches} of queries to alleviate the computational burden of active query generation.
\end{itemize}

\section{Methods}\label{sec:methods}
In this section, we briefly review the methods included in APReL to solve each of the three parts of the problem.

\subsection{Learning from Demonstrations and Preferences}
\textcite{biyik2021learning} proved that it is optimal to initialize the reward prior with demonstrations and then shift to actively collected preference data to update the posterior. Hence, we start with the demonstrations $\mathcal{D}$, which are collected offline, to generate a prior belief over the true reward weights $\omega$:
\begin{align}
	\rho^0(\omega) \propto P(\mathcal{D} \mid \omega) P(\omega) = P(\omega) \prod_{\xi^D \in \mathcal{D}} P(\xi^D \mid \omega)\:,
\end{align}
where the assumption is that these demonstrations are conditionally independent. The probability of a demonstration, i.e., $P(\xi^D \mid \omega)$, is modeled with a computational user model, selected by the system designer.

Then, the belief distribution is updated with proactively generated preference/ranking queries. We first look at how we update the model given these query responses. We will consider how to generate the queries in the next subsection.

The robot asks a new query at each iteration $i$, starting from $i = 0$. We denote the $i^{\text{th}}$ query to be $Q_i$, and the human response to that query $q_i$. Then, using Bayes' theorem and again with the conditional independence assumption,
\begin{align}
	\rho^{i+1}(\omega) \propto P(q_i \mid Q_i, \omega) \cdot \rho^{i}(\omega)\:,
\end{align}
which again requires a computational human response model for $P(q_i \mid Q_i, \omega)$. APReL allows implementing and experimenting with different human models. After learning from the query responses, the posterior belief keeps a distribution for the reward function (or more generally, the human model), which can be used to optimize the robot's behavior.

APReL allows users the option to provide demonstrations to initialize our belief, or to learn solely from preferences, without any provided demonstrations, in which case $\rho^0(\omega)=P(\omega)$. Having presented the way the initial belief is generated from the demonstrations and updated with the human responses to the queries, we now proceed to the second problem to consider how to actively generate the queries.

\subsection{Actively Generating Queries}
The goal of the robot is to generate queries that accurately and efficiently update its estimate of $\omega$, so it should minimize the number of questions the user must answer. 

In addition to allowing the implementation of new acquisition functions, APReL readily provides multiple different options for this active learning problem: volume removal, disagreement, information gain, regret and Thompson sampling. We would ideally find the best \emph{adaptive sequence} of queries for the robot to get accurate, fine-grained information about $\omega$. However, since adaptively reasoning about a sequence of queries is an NP-hard problem \cite{ailon2012active}, the techniques in the literature proceed in a greedy fashion: at each iteration $i$, we choose $Q_i$ while considering only the next posterior $\rho^{i+1}$.

\smallskip
\noindent\textbf{Volume Removal.} Volume removal is a strategy for selecting queries that maximize the expected difference between the prior and the \emph{unnormalized} posterior \cite{sadigh2017active,biyik2019green}. Intuitively, it searches queries $Q_i$ where each possible answer $q_i$ is equally likely given our current belief over $\omega$, which corresponds to the queries where we are most uncertain about which behavior the human will prefer.

\smallskip
\noindent\textbf{Disagreement.} By trying to generate uncertain queries for the human, volume removal formulation often presents queries with very similar trajectories. To overcome this, \textcite{katz2019learning} proposed a multi-objective approach based on a disagreement metric, particularly for pairwise comparison queries. In that approach, the goal is to select two $\omega$'s to maximize both their likelihood and disagreement. The optimal $\omega$'s are then given to a planner, e.g., a reinforcement learning algorithm, which gives the optimal trajectories with respect to those $\omega$'s. These trajectories form the optimal disagreement query. While APReL handles the intermediate planning problem by selecting the trajectories from a predefined trajectory set, its modular structure allows users to implement their own planners.

\smallskip
\noindent\textbf{Mutual Information.} Motivated by the same issue of trajectories being too similar with volume removal, \textcite{biyik2019asking} identified failure cases and proposed maximizing the mutual information between the user response and the reward function instead, which they showed to yield both informative and easy-to-answer questions.

In addition, they showed the learning efficiency can be improved if the users are allowed to respond ``About Equal" to the pairwise comparison queries (called \emph{weak} comparison queries) and these responses are also utilized for learning. APReL also allows having weak comparison queries.

\smallskip
\noindent\textbf{Regret.} For a slightly different formulation, where the goal is to find the optimal behavior rather than learning the reward function, \textcite{wilde2020active} proposed using pairwise comparison queries that maximize a measure of regret between the $\omega$'s that lead to the trajectories in the query. Similar to the disagreement formulation, this acquisition function first optimizes $\omega$'s and then uses a planner to get the trajectories in the query.

\smallskip
\noindent\textbf{Thompson Sampling.} Finally for a similar modified problem, \textcite{tucker2020preference} proposed using a Thompson sampling approach so that the regret will be minimized during the querying process, which implies the optimal trajectory will be quickly learned.

\subsection{Generating Batches of Queries}
When we are fine-tuning our estimate of $\omega$, we want to do so in a way that minimizes the time spent by the human expert. When we generate a new query in between each user response, we spend a significant amount of time to generate the next query. A more time-efficient option is to generate batches of queries at once, so that the time cost will reduce.

APReL has a variety of options for which batch method to use if we are generating batches of queries. We define $b$ to be the number of queries in our batch, and briefly go over each method. First four methods were proposed by \textcite{biyik2018batch}, and the DPP-based method by their follow-up work \cite{biyik2019batch}.

\smallskip
\noindent\textbf{Greedy.} The top $b$ queries that individually optimize the acquisition function form batch. Though computationally efficient, this causes redundancy as we previously discussed.

Therefore, the other batch methods try to increase the diversity in the queries. For this, they first take the top $B$ queries with respect to the acquisition function (so that non-informative queries will not be selected). Then, they try to select $b$ queries out of this reduced set that maximize the total dissimilarity between them with respect to a given distance measure between the queries.

\smallskip
\noindent\textbf{Medoids.} This method partitions the reduced query set into $b$ clusters based on the given distance measure using $k$-medoids algorithm \cite{park2009simple,bauckhage2015numpy}. Then the medoid of each cluster (similar to centroids in $k$-means algorithm) is taken into the batch.

\smallskip
\noindent\textbf{Boundary Medoids.} This method further improves the diversity of the medoids algorithm by applying $k$-medoids only on the boundary of the convex hull of the reduced set. We note that efficiently finding the boundary of the convex hull may not always be possible. Hence, this method works only when the queries can be mapped into an Euclidean space, as in \cite{biyik2018batch}.

\smallskip
\noindent\textbf{Successive Elimination.} The successive elimination algorithm starts with the reduced set and iteratively removes the queries until $b$ of them are left, which are taken as the batch. The removal rule is as follows. It first finds the pair of queries with the shortest distance between them. It then removes the one with the worse acquisition function value out of this pair. Among the heuristic batch selection methods, this often gives the highest data-efficiency.

\smallskip
\noindent\textbf{DPP.} Finally, the DPP method builds a $k$-DPP \cite{kulesza2012determinantal} given the distance between the queries and their acquisition function values. The (approximate) mode of this $k$-DPP distribution, which optimally maximizes the diversity and the informativeness, is then taken as the batch.

\section{APReL Structure and Usage}\label{sec:usage}
\begin{figure}[t]
\includegraphics[width=\columnwidth]{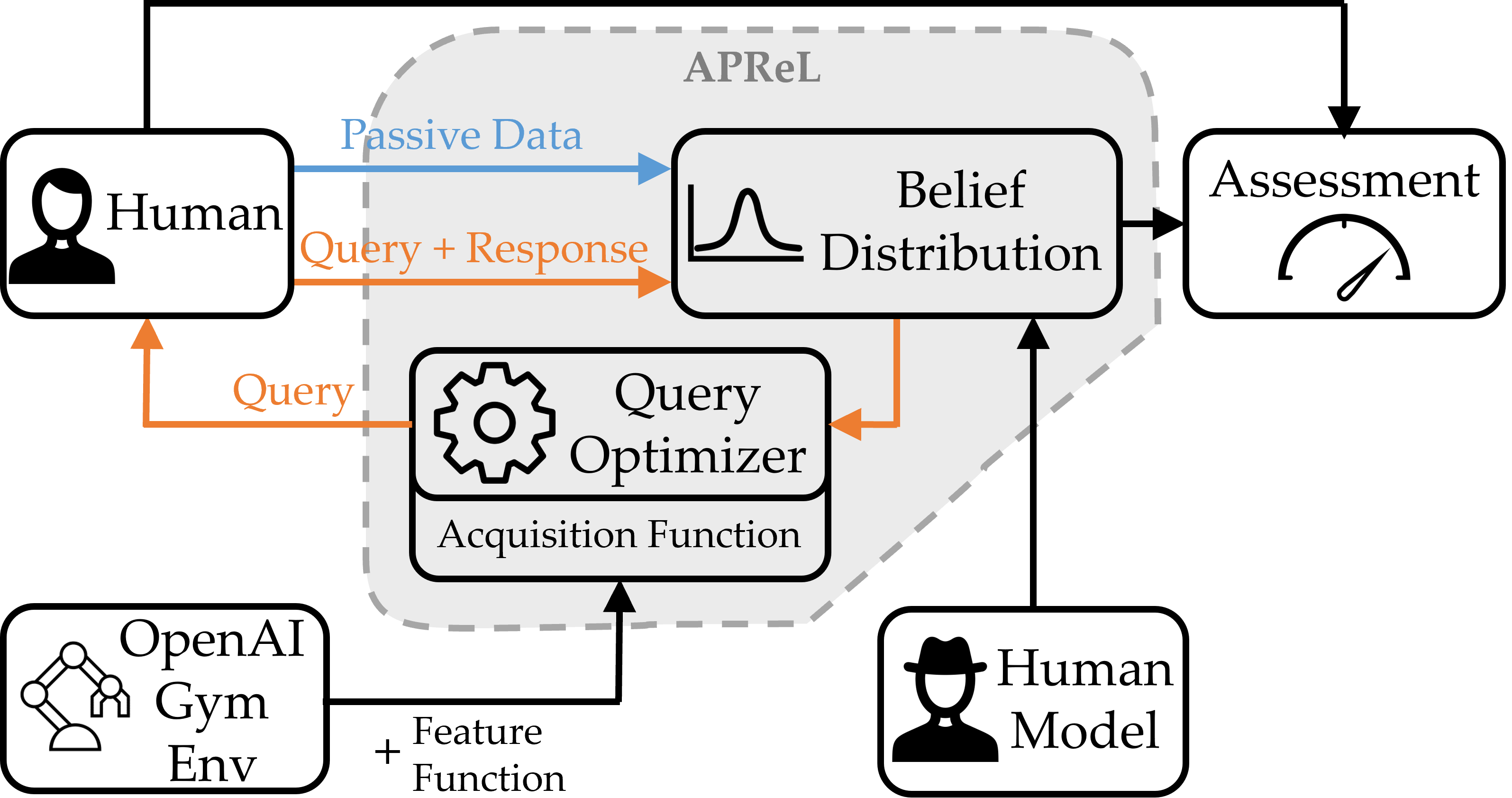}
\centering
\caption{\small APReL assumes a human model for how the human provides demonstrations and responds to the queries. Optionally, some passive data from the actual human, e.g., demonstrations, are used to initialize a belief distribution. Based on this belief, query optimizer then outputs a query that will give the most information about actual human. This query is asked to the human, and their response is used to update the belief, which completes the active learning loop (orange arrows). Finally, the quality of the learned model is assessed in comparison with the actual human.}
\vspace{-10px}
\label{fig:workflow}
\end{figure}

APReL implements multiple modules to provide the researchers with \emph{easy experimentation}. These modules include: query types, user models, belief distributions, query optimizers and different acquisition functions. An overview of APReL's general workflow is shown in Figure~\ref{fig:workflow}. We next briefly go over the modules.

\smallskip
\noindent\textbf{Basics.} APReL implements \emph{Environment} and \emph{Trajectory} classes. An APReL environment requires an OpenAI Gym environment and a features function that maps a given sequence of state-action pairs to a vector of trajectory features as in Eq.~\eqref{eq:reward}. \emph{Trajectory} instances then keep trajectories of the \emph{Environment} along with their features.

\smallskip
\noindent\textbf{Query Types.} As discussed earlier, researchers developed and used several comparison and ranking query types. Among those, APReL readily implements preference queries, weak comparison queries, and full ranking queries. More importantly, the module for query types is customizable, allowing researchers to implement other query types and information sources. As an example, demonstrations are already included in APReL.

\smallskip
\noindent\textbf{User Models.} Preference-based reward learning techniques rely on a human response model, e.g. the softmax model, which gives the probabilities for possible responses conditioned on the query and the reward function. APReL allows to adopt any parametric human model and specify which parameters will be fixed or learned.

\smallskip
\noindent\textbf{Belief Distributions.} After receiving feedback from the human (Human in Fig.~\ref{fig:workflow}), Bayesian learning is performed based on an assumed human model (Human-hat in Fig.~\ref{fig:workflow}) by a belief distribution module. APReL implements the sampling-based posterior distribution model that has been widely employed by the researchers. However, its modular structure also allows to implement other belief distributions, e.g., Gaussian processes.

\smallskip
\noindent\textbf{Query Optimizers.} After updating the belief distribution with the user feedback, a query optimizer completes the active learning loop by optimizing an acquisition function to find the best query to the human. APReL implements the widely-used ``optimize-over-a-trajectory-set" idea for this optimization, and allows the acquisition functions that we discussed earlier. Besides, the optimizer module also implements the batch optimization methods that output a batch of queries using different techniques. All of these three components (optimizer, acquisition functions, batch generator) can be extended to other techniques.

\smallskip
\noindent\textbf{Assessing.} After (or during) learning, it is often desired to assess the quality of the learned reward function or user model. The final module does this by comparing the learned model with the information from the human.

\subsection{Example}
An example runner script is included in APReL that explains how to use the library and the available options. We refer the readers to \url{https://github.com/Stanford-ILIAD/APReL} for installation and running instructions.

\section{Conclusion}\label{sec:conclusion}
In this paper, we introduced our \emph{novel} Python library APReL, which provides a modular framework for experimenting with and developing active preference-based reward learning algorithms. In short term, we are planning to implement more techniques and features, as we discussed in Section~\ref{sec:related_work}. By accepting contributions from the community, our goal is to keep APReL up-to-date with state-of-the-art methods, and possibly extend it to non-Bayesian learning methods.

\clearpage
\AtNextBibliography{\small}
\printbibliography

\end{document}